\documentclass[letterpaper]{article} 
\usepackage[preprint]{aaai2027}  
\usepackage[hyphens]{url}  
\usepackage{graphicx} 
\usepackage{natbib}  
\usepackage{caption} 
\usepackage{algorithm}
\usepackage{algorithmic}
\usepackage{booktabs}
\usepackage{amsmath}
\usepackage{amssymb}
\usepackage{multirow}
\usepackage{subcaption}

\newcommand{\OURMODEL}{TreeAdapter}

\title{\OURMODEL{}: Hierarchical Taxonomy-Guided Adapter Composition\\for Fine-Grained Species Image Generation}

\author{
    Yuze Sun\textsuperscript{1},
    Zhongjie Duan\textsuperscript{1} and
    Yingda Chen\textsuperscript{1}
}

\affiliations{
    \textsuperscript{1}ModelScope Team, Alibaba Group\\
}

\begin{document}

\maketitle

\begin{abstract}
Although general text-to-image models excel in open-domain generation, their performance degrades significantly in specialized downstream domains, particularly when generating images of rare biological species. Hindered by long-tailed distributions, general models struggle to capture subtle fine-grained details, while per-species fine-tuning methods over-isolate individual species and consequently ignore the shared visual features among closely related taxa. To address this, we propose \OURMODEL{}, a novel framework that explicitly leverages hierarchical taxonomic data. Rather than using a monolithic model or independent per-species modules, \OURMODEL{} attaches lightweight adapters to every node of the taxonomic tree. Specifically, leaf-node adapters capture species-specific visual traits, while internal-node adapters encapsulate shared semantics among descendant taxa. We introduce a two-stage training paradigm where ancestor adapters are optimized to model only the residual visual features unexplained by their descendants. This model architecture and training paradigm enable the model to fully leverage hierarchical information, ensuring the accurate generation of visual features for each species. Extensive experiments across three large-scale biodiversity benchmarks demonstrate that \OURMODEL{} achieves state-of-the-art fine-grained generation quality, outperforming both general-purpose and domain-specific baselines. All resources will be open sourced\footnote{https://github.com/modelscope/DiffSynth-Studio}\footnote{https://www.modelscope.cn/models/DiffSynth-Studio/TreeAdapter-KleinBase4B}, including code and models.
\end{abstract}

\section{Introduction}

\begin{figure}[t]
\centering
\includegraphics[width=\columnwidth]{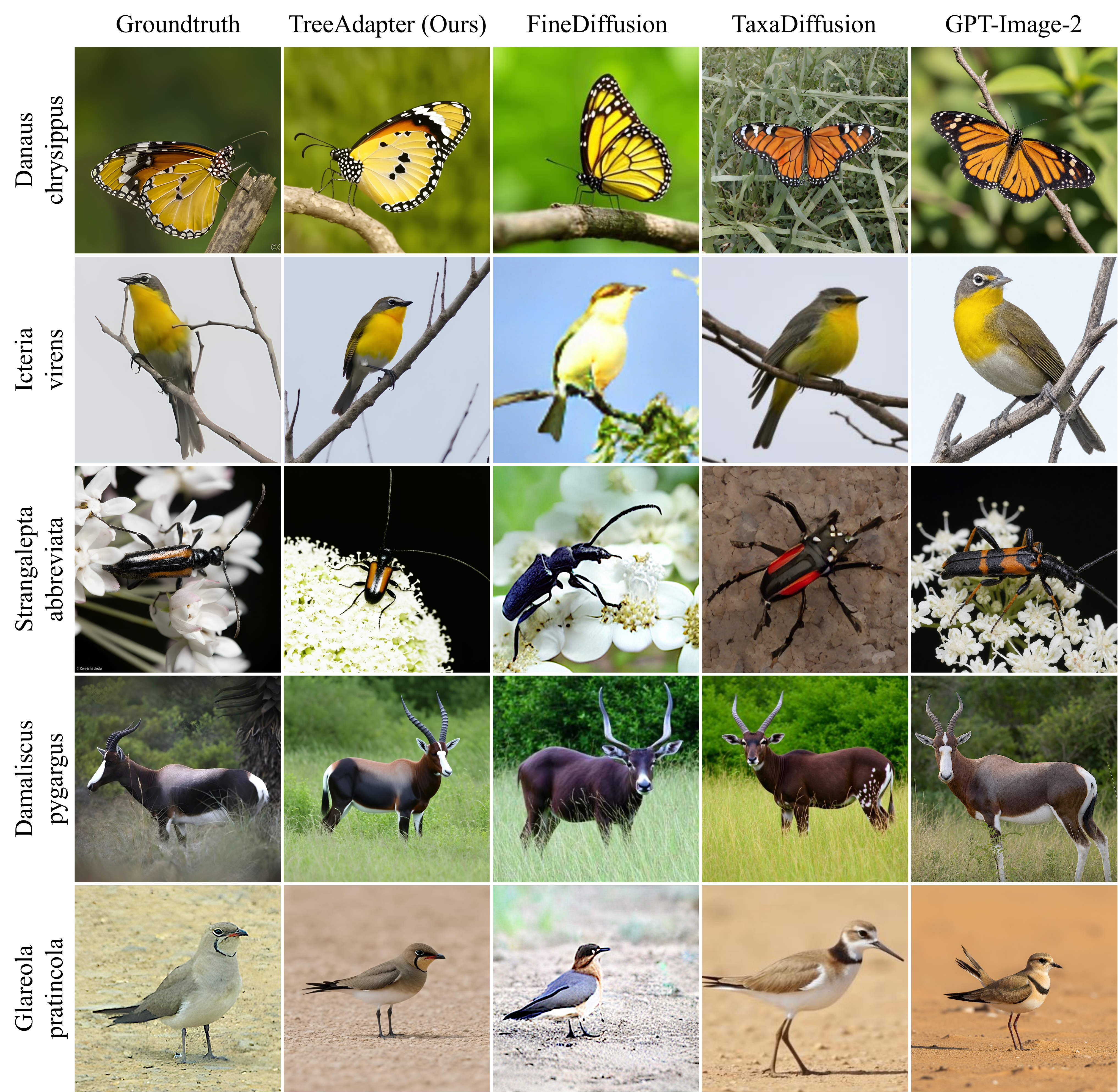}
\caption{\OURMODEL{} generates fine-grained species more faithfully than both general base models and domain-specific methods.}
\label{fig:teaser}
\end{figure}

Text-to-image diffusion models now produce photorealistic images from natural language across a broad range of everyday concepts \cite{rombach2022high,podell2024sdxl,esser2024scaling,flux2024}. Behind this progress lies large-scale web data in which common concepts appear millions of times. Biological species do not fit this regime: they follow a long-tailed distribution in which a handful of charismatic species are abundant while the vast majority are rare, and generating them faithfully would directly support biodiversity documentation, ecological education, and data augmentation for automated recognition systems \cite{van2018inaturalist}. Fine-grained species generation is therefore an important but underserved setting for these models.

Two properties make this setting hard. First, the data is long-tailed, so most species contribute little signal and their appearance is easily overwhelmed by the common categories a model has seen most often. Second, the differences that separate closely related species are visually subtle, a shift in wing venation or a change in coloration boundary, yet these are precisely the traits of central interest to biological research. As a result, when a general diffusion model is prompted for a specific species, it tends to fall back on the coarse category it has seen often and produce a generic image. 

Recent work adapts diffusion models to this setting. FineDiffusion \cite{pan2025finediffusion} attaches lightweight per-category adapters to a shared backbone, and TaxaDiffusion \cite{monsefi2025taxadiffusion} conditions generation on hierarchical taxonomic text embeddings. These methods share a common recipe, giving a pretrained backbone a small amount of extra, category-aware capacity, and it helps. But their generations remain imprecise on the very fine-grained traits that matter, and we attribute this to two compounding gaps. First, they never allocate enough dedicated capacity to any single species: the per-category budget is a small adapter or a handful of shared embeddings, too little to pin down a species' distinctive appearance. Second, they never model the taxonomic hierarchy at a fine enough granularity: taxonomy enters only as coarse global conditioning or is ignored entirely, so the appearance that related species genuinely share is never captured by a dedicated hierarchical module and must instead be relearned separately for each species.

Crucially, biodiversity data does not arrive as a flat list of species. Every image is labeled with a full biological taxonomy that organizes species into nested groups, from broad clades down to individual species. Species grouped together tend to look alike, so this hierarchy is a ready-made map of which species should share visual information. This suggests a \emph{compositional} view: rather than force every species to relearn its group's shared appearance, we can factor that shared signal into a module attached to the group and reuse it across all its members, while a per-species module handles what is unique.

We present \OURMODEL, a hierarchical multi-adapter model system that places a lightweight adapter at every node of the biological taxonomy and, for any target species, composes the adapters along its root-to-leaf path into the diffusion backbone. The design directly fills the two gaps above. To give every species enough dedicated capacity, \OURMODEL{} trains a \emph{species adapter} at each leaf, so that a species is represented by its own parameters and its distinctive traits are captured rather than smoothed into a coarse category. To model the hierarchy at the right granularity, \OURMODEL{} trains an \emph{ancestor adapter} at \emph{each} internal node, shared by exactly the species that node groups, so the appearance common to a clade is captured once and reused across its members. A two-stage training scheme keeps the two roles from competing: species adapters are trained first and then frozen, after which each ancestor adapter is trained on top and only has to model the residual its descendants leave unexplained, namely the genuinely cross-species signal. This keeps ancestor adapters small and lets each species draw on shared structure at every level of the tree at once.

Our key contributions are:

\textbf{Technique.} We present \OURMODEL, a model system that places a lightweight adapter at every node of a biological taxonomy, a species adapter at each leaf and a shared ancestor adapter at each internal node, and composes the adapters along a species' root-to-leaf path on demand, turning the taxonomy into an explicit mechanism for sharing visual information across related species.

\textbf{Method.} We introduce a two-stage training scheme that assigns the two adapter roles cleanly complementary jobs: species adapters are trained first and frozen, so each ancestor adapter need only capture the cross-species signal its descendants leave unexplained. This keeps shared adapters small and makes their contribution a residual rather than a competitor to per-species detail.

\textbf{Effect.} On three large-scale biodiversity benchmarks, \OURMODEL{} achieves the best fine-grained generation quality among strong domain-specific baselines and general base models, and controlled studies confirm that dedicated per-species capacity is necessary and that the shared and species adapters occupy separated, complementary roles, at a far lower per-species parameter cost than enlarging a single flat adapter.

\begin{figure*}[t]
\centering
\includegraphics[width=0.98\textwidth]{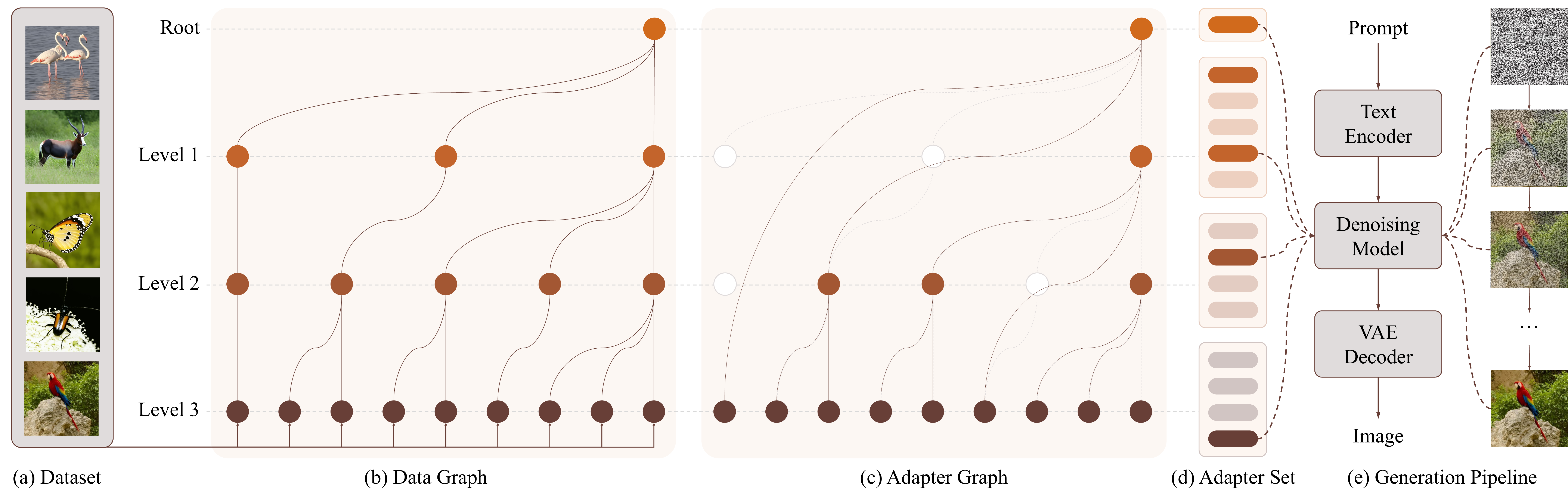}
\caption{Overview of \OURMODEL. \textbf{(a, b)}~Linnaean labels induce a \emph{data graph}: a tree whose internal nodes are taxonomic groups and whose leaves are species. \textbf{(c)}~We prune single-child internal nodes (hollow) and assign an ancestor adapter per internal node and a species adapter per leaf. \textbf{(d)}~The trainable modules consists of a pool of adapters organized by level and shared across subtrees. \textbf{(e)}~At inference we additively compose the species adapter and all ancestor adapters on the target's root-to-leaf path into the denoising model.}
\label{fig:framework}
\end{figure*}

\section{Related Work}

\subsection{Text-to-Image Diffusion Models}

Diffusion models \cite{ho2020denoising,song2020score,song2020denoising} dominate high-fidelity image generation. Latent diffusion \cite{rombach2022high,esser2021taming,van2017neural} works in a compressed latent space, with later work scaling resolution \cite{podell2024sdxl} and adopting rectified-flow transformers \cite{lipman2022flow,esser2024scaling,liu2022flow}; FLUX \cite{labs2025flux1kontextflowmatching,flux2024} pairs a Diffusion Transformer \cite{peebles2023scalable} with flow matching for state-of-the-art quality. These general-purpose models nonetheless lack the specialized knowledge that fine-grained domains require.

\subsection{Parameter-Efficient Fine-Tuning}

Adapter-based fine-tuning adds a small number of trainable parameters to a frozen backbone \cite{zhang2023adding,houlsby2019parameter,mou2024t2i}; Low-Rank Adaptation (LoRA) \cite{hu2022lora}, which we use to realize our adapters, injects a low-rank update $\Delta W = BA$ and is widely used for diffusion customization \cite{ruiz2023dreambooth,kumari2023multi,gal2022image}. Recent work composes several such adapters, for example merging task adapters \cite{yadav2023ties,ilharco2022editing,wortsman2022model} or combining character adapters for multi-subject generation \cite{gu2023mix}. We instead organize adapters along a \emph{containment hierarchy} of nested species groups and train them in two stages, so shared adapters capture only the residual common to a group rather than competing with the per-concept ones.

\subsection{Fine-Grained Image Generation}

Fine-grained biological generation has received limited attention. FineDiffusion \cite{pan2025finediffusion} adds per-category lightweight adapters to a shared backbone; TaxaDiffusion \cite{monsefi2025taxadiffusion} conditions on hierarchical taxonomic text embeddings; EFDiT \cite{wang2025efdit} feeds a tiered super- and child-class embedder into a diffusion transformer to reduce semantic entanglement; and TaxaAdapter \cite{khurana2026taxaadapter} injects vision-taxonomy-model embeddings into a frozen backbone to sharpen species identity. All steer generation through shared class- or taxonomy-conditioned mechanisms with very small per-species budgets, which we find insufficient to separate closely related species. \OURMODEL{} instead pairs a dedicated per-species adapter with shared ancestor adapters, giving enough per-species capacity while amortizing the information common to relatives.

\subsection{Biological Image Datasets}

Large-scale biodiversity datasets ground fine-grained species modeling: iNaturalist \cite{van2018inaturalist} (millions of community observations across diverse taxa), BIOSCAN-1M \cite{gharaee2023step} (standardized dorsal arthropod specimens with DNA barcodes), and FishNet \cite{khan2023fishnet} (curated fish imagery across thousands of species). All organize images by Linnaean taxonomy, providing the hierarchical supervision our approach uses.

\section{Method}

We first formalize fine-grained species generation over hierarchically labeled data, then describe the three components of \OURMODEL: (i) an adapter tree built from the label hierarchy, (ii) a two-stage training scheme that assigns the adapters complementary roles, and (iii) compositional inference that assembles, for any target, the adapters on its path.

\subsection{Problem Formulation}
\label{sec:formulation}

Let $\mathcal{D}=\{(x_i, y_i)\}_{i=1}^{N}$ be a dataset of images $x_i$, each annotated with a \emph{hierarchical label} $y_i$. Unlike a flat class label, $y_i$ is a path of nested groups from coarse to fine,
\begin{equation}
y_i = \big(\ell_1(x_i),\, \ell_2(x_i),\, \ldots,\, \ell_{H}(x_i)\big),
\end{equation}
where $\ell_h$ names the group containing $x_i$ at level $h$, deeper levels refine shallower ones, and the finest level $\ell_H$ is the species. In our datasets the levels instantiate the Linnaean ranks (kingdom, phylum, $\ldots$, species), but the formulation only assumes a nested grouping, not a fixed number or meaning of levels. The nesting makes the label space a rooted tree $\mathcal{T}_0$: each node $v$ is a group, its children partition it into finer groups, and every leaf is one species $s\in\mathcal{S}$.

Given a target species $s$ (or, more generally, any group), the goal is to sample images faithful to that species from a pretrained diffusion backbone with frozen weights $W$. Rather than learn one conditional model for all of $\mathcal{S}$, we allocate capacity over $\mathcal{T}_0$: we attach a trainable adapter to each node and, for a target, activate only the adapters on its root-to-leaf path. This is a form of conditional computation, as in mixture-of-experts models, where only a small, input-dependent subset of parameters is active per sample \cite{shazeer2017outrageously,fedus2022switch,riquelme2021scaling}. In our formulation, the ``experts'' and their routing are fixed by the taxonomy rather than learned: a sample of species $s$ activates exactly the adapters on $s$'s path, at most $H$ of them.

\subsection{Adapter Tree Construction}

\paragraph{Pruning.}
A node of $\mathcal{T}_0$ with a single child groups exactly the same species as that child, so giving it its own adapter would only duplicate the child's, at extra cost. We prune such nodes: while any non-leaf node $v$ has exactly one child, we contract the edge, connecting $v$'s parent directly to its child (Algorithm~\ref{alg:build}); we also drop a root whose single child covers all species, as it would encode nothing beyond the backbone. Let $\mathcal{T}=(\mathcal{V},\mathcal{E})$ denote the pruned tree with node set $\mathcal{V}$. Pruning removes only redundant nodes and never changes which species a surviving node groups; consequently the depth of a species varies across the tree, and a level index no longer corresponds to a fixed taxonomic rank.

\paragraph{Adapter-to-node assignment.}
We attach one adapter to every surviving node through a bijection $\phi:\mathcal{V}\to\{1,\ldots,|\mathcal{V}|\}$ that indexes the adapters. The nodes split into internal nodes $\mathcal{V}_{\mathrm{int}}$ and leaves $\mathcal{S}$, giving two adapter roles:
\begin{itemize}
\item a \emph{species adapter} $\theta_s$ at each leaf $s\in\mathcal{S}$, dedicated to one species' appearance;
\item an \emph{ancestor adapter} $\theta_a$ at each internal node $a\in\mathcal{V}_{\mathrm{int}}$, shared by all species in the subtree rooted at $a$.
\end{itemize}
For a species $s$, let $\pi(s)=(a_1,\ldots,a_{k_s},s)$ be the sequence of nodes on its root-to-leaf path in $\mathcal{T}$, with $a_1$ nearest the root. Because $\phi$ is a bijection, recovering the adapters to use for $s$ is just reading off $\pi(s)$ (Algorithm~\ref{alg:build}, path-lookup step); the same set is used for both training and inference, so every species is generated with the adapters it was trained with.

\begin{algorithm}[t]
\caption{Adapter Tree Construction and Path Lookup}
\label{alg:build}
\begin{algorithmic}[1]
\REQUIRE raw label tree $\mathcal{T}_0$; target species $s$
\ENSURE pruned tree $\mathcal{T}$, adapter index $\phi$, path adapters $L$
\STATE \textbf{// Prune redundant single-child groupings}
\REPEAT
  \STATE $changed \leftarrow \textbf{false}$
  \FOR{each non-leaf node $v \in \mathcal{T}_0$}
    \IF{$v$ has exactly one child $u$}
      \STATE contract edge $(v,u)$: link $\mathrm{parent}(v)$ to $u$
      \STATE $changed \leftarrow \textbf{true}$
    \ENDIF
  \ENDFOR
\UNTIL{$changed = \textbf{false}$}
\STATE $\mathcal{T} \leftarrow \mathcal{T}_0$
\STATE \textbf{// Assign one adapter index per surviving node}
\STATE $i \leftarrow 0$
\FOR{each node $v \in \mathcal{V}$ in BFS order}
  \STATE $\phi(v) \leftarrow i$; \quad $i \leftarrow i + 1$
\ENDFOR
\STATE \textbf{// Look up adapters on the target's path}
\STATE $L \leftarrow [\,]$; \quad $v \leftarrow \mathrm{root}(\mathcal{T})$
\WHILE{$v \neq s$}
  \STATE $v \leftarrow$ child of $v$ on the path to $s$
  \STATE append $\phi(v)$ to $L$
\ENDWHILE
\RETURN $\mathcal{T}, \phi, L$ \COMMENT{$L$: ancestors first, species last}
\end{algorithmic}
\end{algorithm}

\subsection{Two-Stage Hierarchical Training}

Each adapter perturbs the frozen backbone weights $W$ by an additive update $\Delta W(\theta)$, and updates compose by summation. For a target species $s$ with path $\pi(s)$, the effective weights are
\begin{equation}
\label{eq:compose}
W'(s) \;=\; W \;+\; \underbrace{\sum_{j=1}^{k_s}\Delta W(\theta_{a_j})}_{\text{ancestor adapters}} \;+\; \underbrace{\Delta W(\theta_{s})}_{\text{species adapter}} .
\end{equation}
The question is how to train the adapters so that ancestor and species roles do not overlap. Training each adapter independently on its node's images fails for ancestors: an ancestor adapter fit from scratch would re-learn the per-species appearance its species adapters already represent, so the two compete and the shared adapter carries nothing distinct. We therefore train in two stages, illustrated in Figure~\ref{fig:framework}.

\paragraph{Stage 1: species adapters.}
For each species $s$, we train its adapter $\theta_s$ alone on top of the frozen backbone. Writing $W_s = W + \Delta W(\theta_s)$ for the adapted weights, the flow-matching objective is
\begin{equation}
\label{eq:stage1}
\mathcal{L}_1(\theta_s)=\mathbb{E}_{x_0,t,\epsilon}\big[\big\|v_{W_s}(x_t,t,c_s)-(x_0-\epsilon)\big\|^2\big],
\end{equation}
where $x_0$ is an image of $s$, $c_s$ its caption, $x_t$ the flow-matching interpolant at time $t$, and $v_{(\cdot)}$ the velocity prediction of the adapted backbone. Each $\theta_s$ thus specializes fully to one species. As each leaf is optimized independently, Stage~1 parallelizes fully across species with no inter-adapter communication.

\paragraph{Stage 2: ancestor adapters on frozen species adapters.}
We then freeze all species adapters and train the ancestor adapters bottom-up. For an internal node $a$ with descendant species $\mathcal{D}_a$ and an image of species $s\in\mathcal{D}_a$, let
\begin{equation}
\label{eq:stage2-weights}
W_a(s) = W + \Delta W(\theta_a) + \!\!\sum_{b\in\mathcal{F}(s,a)}\!\!\Delta W(\theta_b)
\end{equation}
be the effective weights when only $\theta_a$ is trainable, where $\mathcal{F}(s,a)$ collects the frozen species adapter $\theta_s$ and any ancestor adapters strictly below $a$ on $\pi(s)$. We train $\theta_a$ by
\begin{equation}
\label{eq:stage2}
\min_{\theta_a}\ \mathbb{E}_{s\in\mathcal{D}_a}\,\mathbb{E}_{x_0,t,\epsilon}\big[\big\|v_{W_a(s)}(x_t,t,c_s)-(x_0-\epsilon)\big\|^2\big].
\end{equation}
Since the species-specific appearance is already supplied by the frozen adapters in $\mathcal{F}(s,a)$, the gradient to $\theta_a$ rewards only what they leave unexplained, namely the regularity \emph{common} across $\mathcal{D}_a$. Formally, Stage 2 makes the additive composition of Eq.~\eqref{eq:compose} a \emph{residual decomposition}: $\theta_a$ absorbs the cross-species residual rather than duplicating single-species detail. Two consequences follow: ancestor adapters can be low-rank, because they never re-encode species appearance; and each ancestor's training set is a whole subtree, so its cost is amortized over many species rather than paid once per species.

\subsection{Compositional Inference}

At inference, generating a target species $s$ requires no merging heuristics: we look up $\pi(s)$, sum the low-rank updates of its adapters into the backbone as in Eq.~\eqref{eq:compose}, and sample. Because each ancestor adapter is shared across all species in its subtree, the amortized per-species parameter cost is far below a flat per-species adapter of comparable capacity. In practice we load the backbone once and swap only the small path-specific adapters per species, avoiding redundant backbone loading at scale.

\section{Experiments}

We first evaluate \OURMODEL{} against domain-specific baselines and a strong proprietary model on three large-scale biodiversity datasets. We then run a set of controlled ablations on a self-contained bird subtree that vary the parameter budget, the adapter roles, and the two adapter ranks in turn. We build on FLUX.2-klein \cite{flux2024}, a 4-billion-parameter Diffusion Transformer trained with flow matching.

\subsection{Datasets}

We evaluate on species subsets of three large-scale biodiversity datasets: (1) \textbf{iNaturalist} \cite{van2018inaturalist}: 10{,}000 species across the full animal kingdom ($\sim$4--5 nodes per path after pruning, 2{,}444 ancestor nodes). (2) \textbf{FishNet} \cite{khan2023fishnet}: 9{,}639 ray-finned fish species; sharing the same upper ranks, paths are shorter ($\sim$3--4 nodes, 1{,}861 ancestor nodes). (3) \textbf{BIOSCAN} \cite{gharaee2023step}: 2{,}288 insect species with standardized dorsal specimens ($\sim$3--4 nodes, 495 ancestor nodes). These datasets are severely long-tailed: on FishNet, 3{,}986 species (41.1\%) have a single training image and 8{,}458 (87.7\%) have five or fewer, while on BIOSCAN the
  corresponding counts are 570 species (24.9\%) and 1{,}329 (58.1\%), so the vast majority of species are seen only a handful of times, which is known to be difficult for pretrained diffusion models \cite{samuel2024generating}. In all three, the hierarchical labels $\ell_1,\ldots,\ell_H$ of our formulation are the Linnaean ranks (kingdom down to species). Images are super-resolved by $2\times$ along each side with Real-ESRGAN \cite{wang2021real,wang2018esrgan}: besides sharpening the fine textures that distinguish species, this brings the training images close to the $1024\times1024$ resolution at which mainstream text-to-image backbones are trained, so the model can better exploit its pretrained generative prior. For iNaturalist, whose images span diverse natural scenes, we caption every image with Qwen3.5-9B and prompt with ``a photo of \{common name\} (\{genus species\}). \{caption\}''; FishNet and BIOSCAN are captured in visually uniform settings, so we use ``a photo of \{common name\} (\{genus species\})'' alone. For data splits, iNaturalist trains on official mini train set and tests on one image sampled at random from each species' official validation set; for FishNet and BIOSCAN we first discard species with a single image or incomplete hierarchical labels, then hold out one image per species for testing and train on the rest.

\subsection{Baselines}

We compare against the following methods. \textbf{FLUX.2 Base} is the pretrained FLUX.2-klein-base-4B model\footnote{\url{https://huggingface.co/black-forest-labs/FLUX.2-klein-base}} without fine-tuning, conditioned only on species text prompts. \textbf{FLUX.2 Full FT} fully fine-tunes the backbone on the entire dataset with all species mixed together, and \textbf{FLUX.2 LoRA FT} trains a single rank-16 LoRA on all species jointly (one shared adapter for the whole dataset). \textbf{FineDiffusion} \cite{pan2025finediffusion} uses per-category lightweight adapters with a shared backbone, via their released implementation, and \textbf{TaxaDiffusion} \cite{monsefi2025taxadiffusion} conditions on taxonomy-aware text embeddings with hierarchical class labels. Finally, \textbf{GPT-Image-2} is a strong proprietary text-to-image model prompted with the species name and its taxonomy, included as a qualitative reference point (Figures~\ref{fig:teaser} and~\ref{fig:intragenus}).

\subsection{Evaluation Metrics}

We employ five complementary metrics to comprehensively evaluate generation quality. \textbf{FID} \cite{heusel2017gans} (Fr\'{e}chet Inception Distance) measures distributional similarity between generated and real images, averaged over all test-set images; lower is better. \textbf{LPIPS} \cite{zhang2018unreasonable} (Learned Perceptual Image Patch Similarity) measures perceptual distance to the reference image; lower is better. \textbf{CLIP Score} \cite{radford2021learning} is the cosine similarity between CLIP embeddings of the generated image and text prompt; higher is better. \textbf{BioCLIP Score} \cite{stevens2024bioclip,gu2026bioclip} is a domain-specific CLIP score computed with BioCLIP~2, a vision-language model trained on hundreds of millions of organism images under hierarchical taxonomic supervision; because its representation is aligned with the tree of life rather than generic web text, it separates visually similar species far more reliably than general-purpose CLIP, making it the more faithful measure of whether a generated image preserves species-level identity (higher is better). Finally, \textbf{Aesthetic Score} predicts aesthetic quality using a CLIP-based aesthetic predictor; higher is better.

\subsection{Implementation Details}

Each adapter is realized as a LoRA \cite{hu2022lora} module: for a backbone weight matrix $W\in\mathbb{R}^{d_{\text{out}}\times d_{\text{in}}}$ the update is $\Delta W = BA$ with $B\in\mathbb{R}^{d_{\text{out}}\times r}$, $A\in\mathbb{R}^{r\times d_{\text{in}}}$, and $r\ll d$; the rank $r$ is set independently per adapter, so species and ancestor adapters need not share a rank. All adapters attach to the same attention and MLP projections across the backbone's transformer blocks. By default \OURMODEL{} uses species rank 8 and ancestor rank 4 (the rank-sweep knee, Table~\ref{tab:ablation_rank}), all at resolution $1024\times1024$ with the AdamW optimizer. Stage~1 trains each species adapter for 500 steps at learning rate $3\times10^{-4}$. Stage~2 trains all ancestor adapters for 1 epoch at learning rate $5\times10^{-5}$. At inference we use 40 denoising steps with classifier-free guidance scale 4.0 \cite{ho2022classifier}. On iNaturalist, Stage~1 trains the 10{,}000 species adapters on 16 NVIDIA A100 GPUs in about 60 hours, and Stage~2 trains the 2{,}444 ancestor adapters on 8 A100 GPUs in about 36 hours. All experiments are implemented with the DiffSynth-Studio framework \cite{duan2024diffsynth}.

\subsection{Main Results}

\begin{table*}[t]
\centering
\setlength{\tabcolsep}{7.7pt}
\caption{Quantitative comparison across three biodiversity datasets. $\downarrow$: lower is better. $\uparrow$: higher is better. Best per row in \textbf{bold}, second best \underline{underlined}. \OURMODEL{} attains the best FID, CLIP Score, and BioCLIP on all three datasets; the remaining top scores (LPIPS on iNaturalist, Aesthetic on every dataset) go to the base model or a single shared adapter, which trade per-species identity for smoother, more generic images.}
\label{tab:main_results}
\begin{tabular}{ll|cccccc}
\toprule
Dataset & Metric & Base & Full FT & LoRA FT & FineDiffusion & TaxaDiffusion & \OURMODEL{} (Ours) \\
\midrule
\multirow{5}{*}{iNaturalist}
& FID $\downarrow$ & 30.9 & \underline{19.6} & 20.9 & 22.4 & 61.5 & \textbf{18.8} \\
& LPIPS $\downarrow$ & 0.693 & 0.664 & \textbf{0.649} & 0.808 & 0.735 & \underline{0.655} \\
& CLIP Score $\uparrow$ & \underline{36.9} & 36.5 & 36.6 & 29.3 & 25.2 & \textbf{37.8} \\
& BioCLIP $\uparrow$ & 64.7 & 66.3 & 65.2 & \underline{68.1} & 62.1 & \textbf{71.8} \\
& Aesthetic $\uparrow$ & \textbf{5.25} & 4.90 & 5.01 & 4.46 & 4.81 & \underline{5.07} \\
\midrule
\multirow{5}{*}{FishNet}
& FID $\downarrow$ & 69.1 & 45.9 & 56.9 & \underline{33.3} & 58.8 & \textbf{17.0} \\
& LPIPS $\downarrow$ & 0.873 & 0.850 & 0.873 & \underline{0.795} & 0.825 & \textbf{0.749} \\
& CLIP Score $\uparrow$ & 25.3 & \underline{27.4} & 27.1 & 27.3 & 26.5 & \textbf{28.5} \\
& BioCLIP $\uparrow$ & 58.2 & 65.1 & 60.9 & 66.2 & \underline{66.7} & \textbf{72.6} \\
& Aesthetic $\uparrow$ & \textbf{5.34} & 5.04 & 4.75 & 4.62 & 4.94 & \underline{5.19} \\
\midrule
\multirow{5}{*}{BIOSCAN}
& FID $\downarrow$ & 98.3 & 52.7 & 93.7 & \underline{36.0} & 72.7 & \textbf{20.9} \\
& LPIPS $\downarrow$ & 0.796 & \underline{0.488} & 0.512 & 0.503 & 0.640 & \textbf{0.457} \\
& CLIP Score $\uparrow$ & 24.4 & \underline{27.4} & 26.5 & 27.1 & 25.2 & \textbf{27.5} \\
& BioCLIP $\uparrow$ & 51.6 & 63.4 & 60.5 & \underline{65.5} & 64.1 & \textbf{70.5} \\
& Aesthetic $\uparrow$ & \textbf{5.36} & 4.45 & 4.42 & 4.42 & 4.47 & \underline{4.58} \\
\bottomrule
\end{tabular}
\end{table*}

\begin{figure}[t]
\centering
\includegraphics[width=\columnwidth]{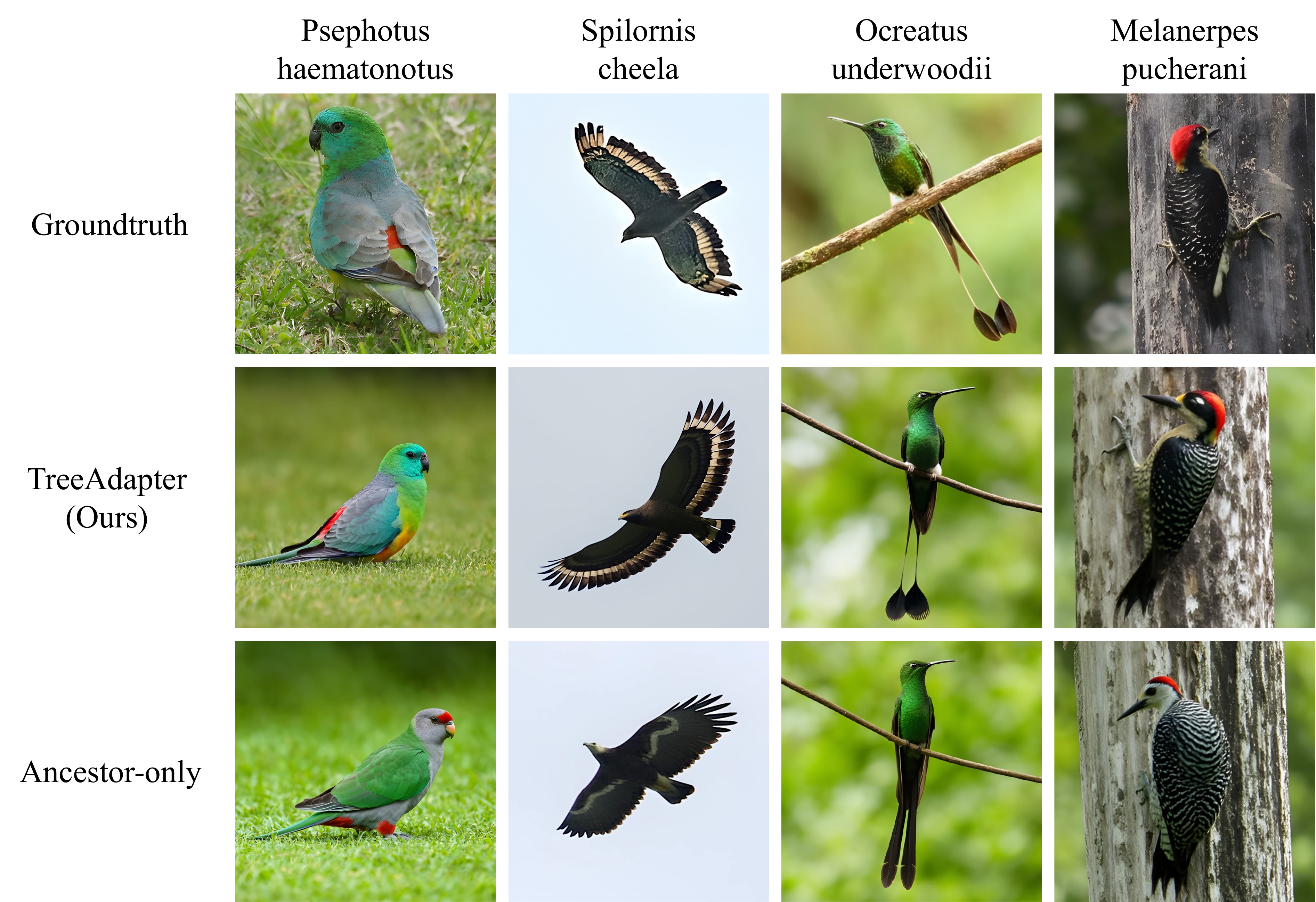}
\caption{A dedicated per-species adapter is mandatory. each column is a different species.}
\label{fig:ancestor}
\end{figure}

\begin{figure}[t]
\centering
\includegraphics[width=0.98\columnwidth]{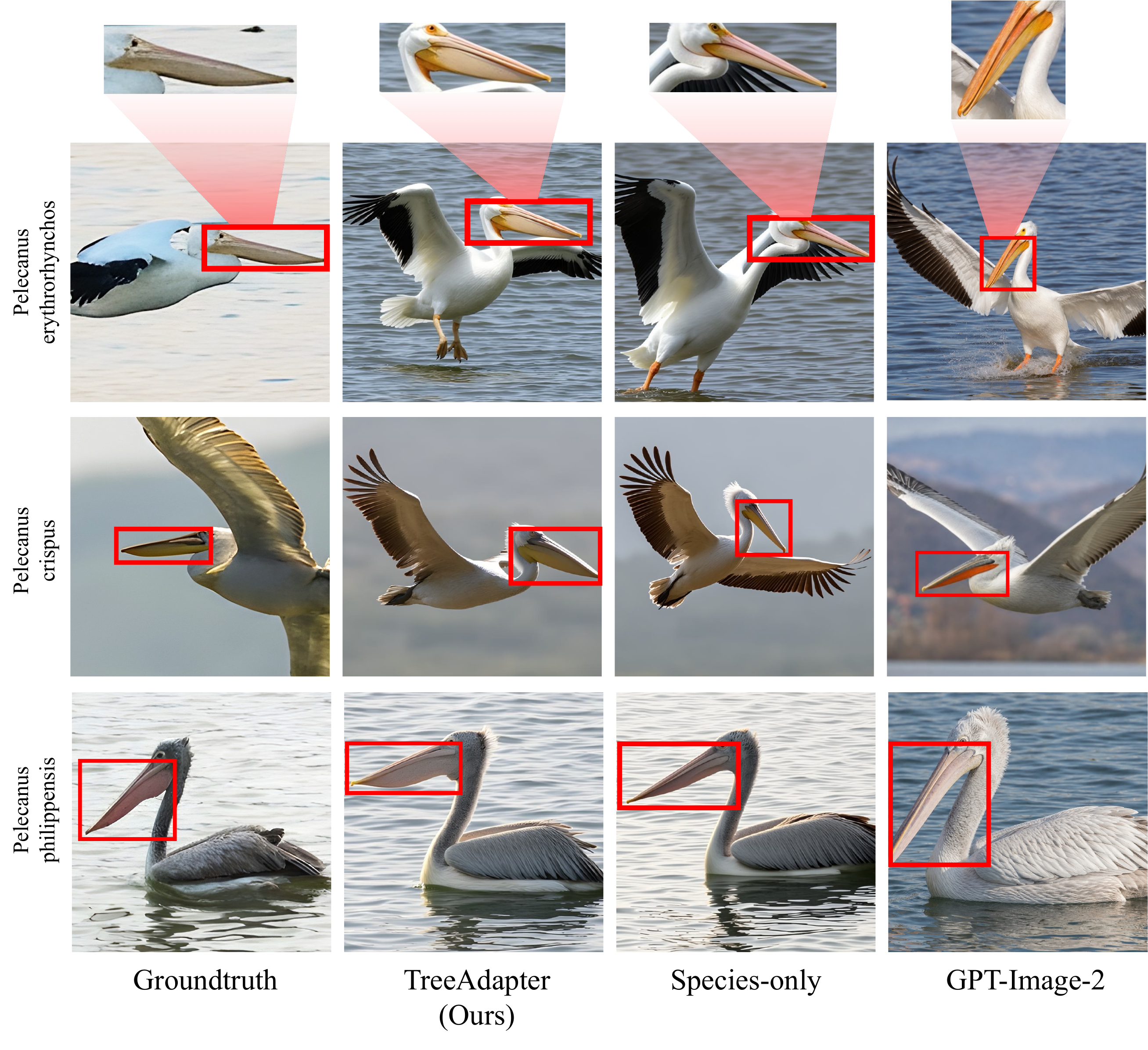}
\caption{Genus \emph{Pelecanus}. Each row is a different species; the red inset zooms the bill-and-pouch, a trait shared across the genus. \OURMODEL{} reproduces the shared bill shape consistently across species, whereas the species-only variant renders it less faithfully.}
\label{fig:intragenus}
\end{figure}

Table~\ref{tab:main_results} reports the full-scale comparison. \OURMODEL{} attains the best FID, CLIP Score, and BioCLIP on all three datasets, including the heavily long-tailed FishNet and BIOSCAN on which most species are seen only a handful of times, the regime that is the norm for biodiversity data.

\textbf{Dedicated per-species capacity is necessary.} A clear gap separates methods that inject only shared or global information from those that spend parameters on each species. The base model, full fine-tuning, and a single shared LoRA all condition generation on group-level signals, whether a prompt or one weight set that must serve every species at once, and FineDiffusion and TaxaDiffusion hit the same ceiling, steering generation through lightweight class- or taxonomy-conditioned mechanisms with tiny per-species budgets. Our ablations make the failure concrete: an ancestor-only model, a shared group-conditioned model with no per-species adapter, reaches good FID but the worst BioCLIP of any variant, rendering closely related species nearly interchangeable and losing their identity (Figure~\ref{fig:ancestor}, Table~\ref{tab:ablation_roles}). Fine-grained generation is thus not a matter of better conditioning on top of a shared model, but of explicitly spending capacity per species.

\textbf{Ancestor adapters capture genuine cross-species structure.} A recurring premise in this line of work is that related species share visual traits worth modeling once. We make that premise concrete and testable in our setting: because the ancestor role is isolated in a dedicated module, we can inspect what it learns. Within a genus, the ancestor adapter reproduces the trait its species share, such as the bill-and-pouch shape across \emph{Pelecanus} (Figure~\ref{fig:intragenus}), consistently across species, whereas a species-only model, lacking any shared module, renders the same trait inconsistently from one species to the next. Our role ablation quantifies the effect: adding the shared ancestor role is what moves FID while leaving per-species identity intact (Table~\ref{tab:ablation_roles}). The shared structure is thus not merely a parameter-saving trick but a real, inspectable signal the hierarchy exposes.

\textbf{Combining the two adapter roles beats scaling one.} Given that per-species capacity is required and that shared structure genuinely exists, the question is how to spend the budget. A flat per-species adapter must relearn the shared structure inside every leaf; \OURMODEL{} instead lets ancestor adapters carry it once. This is far more parameter-efficient: on the bird subtree even the smallest split, ancestor rank~1 with species rank~1, beats the flat rank-16 per-species adapter on both FID and BioCLIP (Table~\ref{tab:ablation_matched}), since each species adapter then only has to encode what is unique to that species. Table~\ref{tab:main_results} shows this efficiency carries to all three datasets, where \OURMODEL{} leads on FID, CLIP, and BioCLIP.

\subsection{Ablation Studies}

To keep ablation compute tractable, we run all ablations on a self-contained subtree of iNaturalist drawn from the bird class \emph{Aves}. Rather than the whole class, which at the iNaturalist scale is dominated by a few very large orders and is structurally unrepresentative, we keep five whole mid-sized orders (Accipitriformes, Pelecaniformes, Piciformes, Caprimulgiformes, Psittaciformes), yielding \textbf{275 species and 79 ancestor nodes} after pruning. Taking \emph{whole} orders leaves no branch thinned, so the within-genus structure the hierarchy exploits stays intact, and the subtree reproduces the branching profile of the full 10{,}000-species tree at every level while the full \emph{Aves} class does not (Appendix~\ref{app:subtree}); conclusions here are thus expected to transfer. The subtree preserves the adapter assignment of the full tree, so the species adapters are exactly those from Stage~1 and only the 79 ancestor adapters are retrained per configuration.

We report FID and BioCLIP, which probe the two complementary aspects of fine-grained fidelity. FID measures how well the \emph{distribution} of generated images matches the real images of a group, i.e.\ group-level realism. BioCLIP scores how accurately a \emph{single} generated image reproduces its target species' identity, i.e.\ per-species correctness. A fine-grained generator must satisfy both, and, as the studies below show, in our model the two are governed by the two different adapter roles. Three studies isolate, in turn, (i) whether hierarchy helps at a matched parameter budget, (ii) whether both adapter roles are necessary, and (iii) how the species and ancestor ranks should be set.

\subsubsection{Hierarchy vs.\ Flat at Matched Budget.}
We write a \OURMODEL{} configuration as a pair $(r_a, r_s)$ of ancestor rank $r_a$ and species rank $r_s$, and pair each against the flat per-species adapter whose rank equals the combined budget $r_a{+}r_s$. Table~\ref{tab:ablation_matched} shows hierarchy wins at every budget: the shared ancestor adapter supplies cross-species structure the flat adapter would otherwise relearn inside every leaf. The effect is starkest at the low end. The smallest split $(1,1)$ already beats the \emph{largest} flat baseline of rank~16 on both FID ($31.09$ vs.\ $32.89$) and BioCLIP ($73.49$ vs.\ $73.36$), matching a rank-16 adapter's discriminative power at a fraction of the per-species cost, since the ancestor rank is shared across all 79 internal nodes.

\begin{table}[t]
\centering
\caption{Hierarchy vs.\ flat at a matched parameter budget (Aves). Each \OURMODEL{} row, written as a pair (ancestor rank, species rank), is paired with the flat per-species adapter of the equivalent total rank. Best of each pair in \textbf{bold}.}
\label{tab:ablation_matched}
\setlength{\tabcolsep}{3pt}
\begin{tabular}{ll|cc|cc}
\toprule
 & & \multicolumn{2}{c|}{\OURMODEL} & \multicolumn{2}{c}{Flat} \\
$(r_a, r_s)$ & Flat rank & FID $\downarrow$ & BioCLIP $\uparrow$ & FID $\downarrow$ & BioCLIP $\uparrow$ \\
\midrule
$(1,1)$ & 2  & \textbf{31.09} & \textbf{73.49} & 34.00 & 73.26 \\
$(2,2)$ & 4  & \textbf{29.98} & \textbf{73.49} & 33.57 & 73.32 \\
$(4,4)$ & 8  & \textbf{28.64} & \textbf{73.57} & 33.42 & 73.38 \\
$(4,8)$ & 16 & \textbf{27.23} & \textbf{73.63} & 32.89 & 73.36 \\
\bottomrule
\end{tabular}
\end{table}

\subsubsection{Isolating Each Adapter Role.}
We compare species-only adapters (the flat baseline, ranks $\{4,8,16\}$), ancestor-only adapters trained from scratch with no species adapters (ranks $\{4,8,16\}$), and our split at $(4,8)$. Table~\ref{tab:ablation_roles} shows a clean division of labour. Ancestor-only attains the \emph{best} FID of any single-role variant ($27.51$) but the \emph{worst} BioCLIP ($70.47$): with no per-species capacity it renders related birds nearly interchangeable, matching the image distribution (low FID) but losing identity (low BioCLIP), as Figure~\ref{fig:ancestor} shows directly. Species-only is the mirror image: it keeps identity (BioCLIP $\approx 73.3$) but, re-encoding the shared structure in every leaf, its FID stays flat near $33$ regardless of rank. Only the split reaches the best FID \emph{and} BioCLIP ($27.23/73.63$), confirming the roles are complementary: the ancestor adapter carries the cross-species signal, the species adapter the per-species identity.

\begin{table}[t]
\centering
\caption{Each adapter role governs a different metric (Aves). Ancestor-only attains the best FID but the worst BioCLIP (shared appearance only, lost identity); species-only is the mirror image (identity kept, distribution fit flat across rank). Only the split wins both. Best in \textbf{bold}.}
\label{tab:ablation_roles}
\begin{tabular}{l|c|c|c}
\toprule
Configuration & rank & FID $\downarrow$ & BioCLIP $\uparrow$ \\
\midrule
Species-only & 4 & 33.57 & 73.32 \\
Species-only & 8 & 33.42 & 73.38 \\
Species-only & 16 & 32.89 & 73.36 \\
Ancestor-only & 4 & 29.07 & 71.28 \\
Ancestor-only & 8 & 28.70 & 71.00 \\
Ancestor-only & 16 & 27.51 & 70.47 \\
\OURMODEL & $(4,8)$ & \textbf{27.23} & \textbf{73.63} \\
\bottomrule
\end{tabular}
\end{table}

\subsubsection{Choosing the Ranks.}
Finally we sweep both ranks (species $\{4,8\}$, ancestor $\{1,2,4\}$, ancestor $\le$ species). Table~\ref{tab:ablation_rank} shows the two act as near-independent knobs, one per metric. \emph{Ancestor rank is the FID lever}: raising it $1\!\to\!4$ lowers FID by 3--4 points ($31.60\!\to\!28.64$ at species 4; $31.16\!\to\!27.23$ at species 8) while BioCLIP barely moves. \emph{Species rank is the BioCLIP lever}: raising it $4\!\to\!8$ improves BioCLIP but leaves FID nearly unchanged, exactly what the residual decomposition predicts. Both effects saturate quickly, so we adopt the $(4,8)$ configuration, the joint knee.

\begin{table}[t]
\centering
\caption{Joint rank sweep on Aves (ancestor rank $\le$ species rank). The two ranks act as near-independent levers: the ancestor rank controls FID, the species rank controls BioCLIP, and both saturate quickly. Best in \textbf{bold}.}
\label{tab:ablation_rank}
\begin{tabular}{cc|cc}
\toprule
Anc.\ rank & Sp.\ rank & FID $\downarrow$ & BioCLIP $\uparrow$ \\
\midrule
1 & 4  & 31.60 & 73.55 \\
2 & 4  & 30.84 & 73.56 \\
4 & 4  & 28.64 & 73.57 \\
1 & 8  & 31.16 & 73.63 \\
2 & 8  & 30.33 & 73.64 \\
4 & 8  & \textbf{27.23} & \textbf{73.63} \\
\bottomrule
\end{tabular}
\end{table}

Together the three studies establish that the hierarchical split drives the gains: it beats a flat adapter of equal budget, and even the smallest $(1,1)$ split beats a flat rank-16 adapter (Table~\ref{tab:ablation_matched}); it needs both adapter roles, an FID-governing ancestor and a BioCLIP-governing species adapter (Table~\ref{tab:ablation_roles}, Figures~\ref{fig:ancestor} and~\ref{fig:intragenus}); and the two map onto near-independent rank knobs, so small ancestor and modest species ranks already reach near-peak quality (Table~\ref{tab:ablation_rank}; Appendix~\ref{app:pareto} shows this geometrically).

\section{Discussion}

\OURMODEL{} also has limitations worth noting. It requires a known nested grouping (here a biological taxonomy), which may be unavailable or noisy in other domains. The two-stage training adds cost for the ancestor adapters, though Stage~1 is fully parallel and each ancestor adapter is small. The additive composition also assumes species and ancestor contributions combine linearly, which may not hold for highly dissimilar taxa sharing superficial visual overlap. Future directions include learning the grouping rather than assuming one, extending to non-biological nested domains such as product categories or architectural styles, and letting a species draw on more than one branch when taxonomically distant species share distinctive appearance traits.

\section{Conclusion}

We presented \OURMODEL, which factors per-species appearance into a dedicated species adapter and shared ancestor adapters, trained in two stages so the shared adapters absorb only the cross-species residual and composed additively at inference. Across three large-scale biodiversity benchmarks \OURMODEL{} achieves the best fine-grained generation quality among strong domain-specific baselines, and controlled studies show that dedicated per-species capacity is necessary and that the ancestor and species adapters occupy separated, complementary roles: the ancestor adapters carry the signal shared within a group, the species adapters each species' distinctive appearance. Fine-grained species generation thus needs dedicated per-species capacity, and organizing it hierarchically spends that capacity far more efficiently.

\bibliography{references}
\clearpage
\appendix
\section{Subtree Structure for Ablations}
\label{app:subtree}
All ablations use \emph{aves\_mini}, five whole mid-sized \emph{Aves} orders (275 species, 79 ancestor nodes after pruning). Table~\ref{tab:subtree_structure} shows why we ablate on this subtree rather than the whole \emph{Aves} class: \emph{aves\_mini} matches the branching profile of the full 10{,}000-species iNaturalist tree at every level, whereas the full \emph{Aves} class does not, since its Class$\rightarrow$Order fan-out is a degenerate $32$-way star, so conclusions on \emph{aves\_mini} are expected to transfer to the full tree.

\begin{table}[t]
\centering
\setlength{\tabcolsep}{4pt}
\caption{Branching statistics: \emph{aves\_mini} tracks the full tree at every level; the full \emph{Aves} class does not.}
\label{tab:subtree_structure}
\begin{tabular}{l|ccc}
\toprule
Branching & Full tree & \emph{Aves} & \emph{aves\_mini} \\
statistic & (10K sp.) & (1.5K sp.) & (275 sp., ours) \\
\midrule
Internal/leaf ratio       & 0.634 & 0.559 & 0.575 \\
Class$\rightarrow$Order    & 5.35  & 32.0  & 5.00  \\
Order$\rightarrow$Family   & 4.04  & 4.59  & 4.00  \\
Genus$\rightarrow$species  & 2.04  & 2.29  & 2.12  \\
\bottomrule
\end{tabular}
\end{table}

\section{Geometric View of the Role Split}
\label{app:pareto}
\begin{figure}[t]
\centering
\begin{subfigure}{0.49\columnwidth}
\centering
\includegraphics[width=\columnwidth]{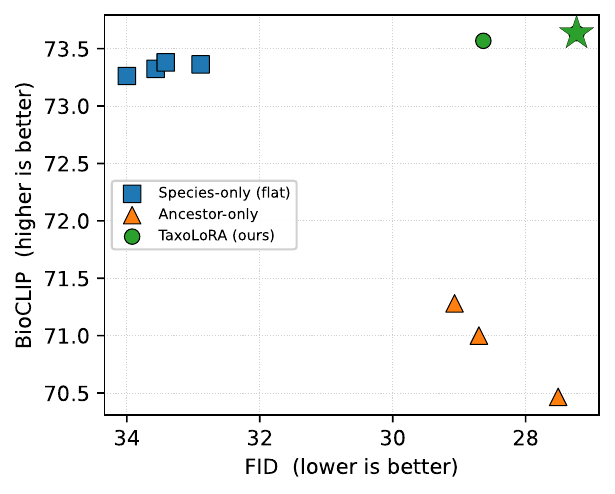}
\caption{Adapter roles}
\label{fig:pareto_roles}
\end{subfigure}
\hfill
\begin{subfigure}{0.49\columnwidth}
\centering
\includegraphics[width=\columnwidth]{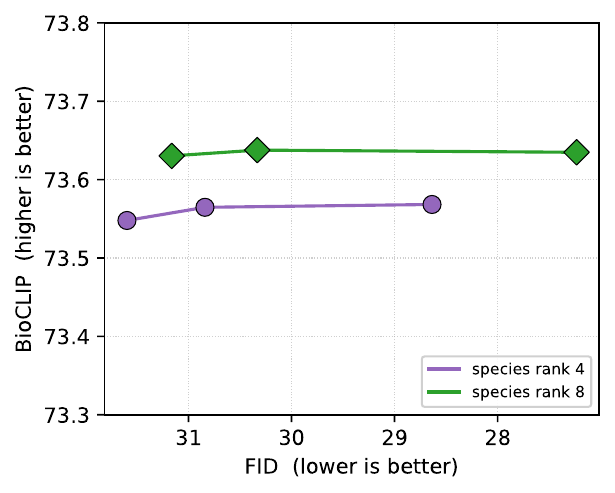}
\caption{Rank knobs}
\label{fig:pareto_ranks}
\end{subfigure}
\caption{The FID--BioCLIP plane (axes oriented so the ideal corner is top-right). \textbf{(a)} The two single-role families occupy opposite failure corners, species-only stuck at high FID and ancestor-only at low BioCLIP, while \OURMODEL{} configurations dominate both (Table~\ref{tab:ablation_roles}). \textbf{(b)} Within the \OURMODEL{} grid the two ranks act as near-independent knobs: raising the ancestor rank lowers FID, stepping from species rank 4 to 8 raises BioCLIP (Table~\ref{tab:ablation_rank}).}
\label{fig:pareto}
\end{figure}

Figure~\ref{fig:pareto} plots the ablation numbers of Tables~\ref{tab:ablation_roles} and~\ref{tab:ablation_rank} on the FID--BioCLIP plane, making the division of labour visible as geometry.

\section{Prompts for Figures}
\label{app:prompts}
Table~\ref{tab:fig_prompts} lists the exact prompt used to generate each species shown in the qualitative figures.

\begin{table*}[t]
\centering
\setlength{\tabcolsep}{4pt}
\caption{Prompts used for the species shown in each qualitative figure.}
\label{tab:fig_prompts}
\begin{tabular}{llp{0.6\textwidth}}
\toprule
Figure & Species (\emph{genus species}) & Prompt \\
\midrule
\multirow{5}{*}{Fig.~\ref{fig:teaser}}
& Danaus chrysippus & a photo of African Monarch (Danaus chrysippus). A vibrant butterfly with yellow wings, black borders, and white spots perches on a twig. The background is softly blurred green, highlighting the butterfly's intricate patterns under natural light. \\
& Icteria virens & a photo of Yellow-breasted Chat (Icteria virens). A small bird with a bright yellow chest and grayish wings perches on a bare branch against a pale sky, its head turned slightly to the side. \\
& Strangalepta abbreviata & a photo of Strangalepta Flower Longhorn (Strangalepta abbreviata). A black and orange beetle perches on white flowers, its long antennae extending outward. The background is dark, highlighting the insect's vibrant colors and delicate floral surroundings. \\
& Damaliscus pygargus & a photo of Bontebok/Blesbok (Damaliscus pygargus). A dark brown antelope with white markings stands in tall grass, its curved horns prominent. The background is lush greenery, softly lit by natural light. \\
& Glareola pratincola & a photo of Collared Pratincole (Glareola pratincola). A small bird with a long tail and short wings stands on sandy ground. Its plumage is light brown above, white below, with a dark collar around its neck. The background is a blurred expanse of sand. \\
\midrule
\multirow{4}{*}{Fig.~\ref{fig:ancestor}}
& Psephotus haematonotus & a photo of Red-rumped Parrot (Psephotus haematonotus). A vibrant green and gray parrot stands on grass, its red rump visible. Soft natural light highlights its feathers against a blurred green backdrop. \\
& Spilornis cheela & a photo of Crested Serpent Eagle (Spilornis cheela). A bird soars gracefully against a pale sky, its wings spread wide, showcasing dark feathers with lighter edges. The scene is serene, with soft lighting highlighting the bird's elegant form. \\
& Ocreatus underwoodii & a photo of Booted Racket-tail (Ocreatus underwoodii). A vibrant green hummingbird perches on a branch, showcasing its long tail feathers with racket-like tips. The background is softly blurred greenery, highlighting the bird's vivid colors and delicate features under natural light. \\
& Melanerpes pucherani & a photo of Black-cheeked Woodpecker (Melanerpes pucherani). A woodpecker with a red crown and black-and-white speckled body clings vertically to a weathered tree trunk, set against a soft-focus green forest backdrop under natural daylight. \\
\midrule
\multirow{3}{*}{Fig.~\ref{fig:intragenus}}
& Pelecanus erythrorhynchos & a photo of American White Pelican (Pelecanus erythrorhynchos). A large white bird with black wingtips spreads its wings over rippling water, orange beak and legs visible, captured mid-action in natural light. \\
& Pelecanus crispus & a photo of Dalmatian Pelican (Pelecanus crispus). A large bird with outstretched wings soars gracefully against a soft, blurred landscape. Its long beak and expansive feathers are highlighted by natural light. \\
& Pelecanus philippensis & a photo of Grey Pelican (Pelecanus philippensis). A grey pelican floats calmly on rippling water, its long beak and feathered neck prominent. Soft sunlight reflects off the gentle waves surrounding it. \\
\bottomrule
\end{tabular}
\end{table*}
These prompts are produced by pairing each species' common and scientific name with a caption of its reference image. For iNaturalist we generate these captions with Qwen3.5-9B using the following instruction, which deliberately withholds the species name so the caption describes only visual context (pose, background, lighting) rather than restating identity:

\begin{quote}
\ttfamily\small
Describe this photograph of \{common\_name\} (scientific name: \{latin\_name\}). Focus on: pose, background and lighting in the frame. Do NOT mention the species name or any taxonomic term in your description. Keep under 25 words. Use plain descriptive English.
\end{quote}
\section{Dataset Storage and Training Cost}
\label{app:cost}
Table~\ref{tab:dataset_cost} reports the total storage footprint and end-to-end training time for each dataset.

\begin{table*}[t]
\centering
\setlength{\tabcolsep}{4pt}
\caption{Total storage footprint and training hours per dataset. Species adapters use LoRA rank 8 and ancestor adapters use LoRA rank 4.}
\label{tab:dataset_cost}
\begin{tabular}{l|ccccc}
\toprule
Dataset & Stage 1 Storage & Stage 1 GPU days & Stage 2 Storage & Stage 2 GPU days & test set size\\
\midrule
iNaturalist & 138\,GB & 40 & 16.8\,GB & 12 & 10000\\
FishNet     & 133\,GB & 40 & 12.8\,GB & 4 & 9639\\
BIOSCAN     & 31.5\,GB & 8 & 3.40\,GB & 3 & 2288\\
\bottomrule
\end{tabular}
\end{table*}
\end{document}